\documentclass[10pt,twocolumn,letterpaper]{article}

\usepackage{wacv}
\usepackage{times}
\usepackage{epsfig}
\usepackage{graphicx,graphics}
\usepackage{amsmath}
\usepackage{amssymb}
\usepackage{textcomp}
\usepackage{xcolor}

\usepackage{balance}

\usepackage{algorithm, algorithmic}

\usepackage{color}
\definecolor{red}{rgb}{1.00,0.00,0.00}
\definecolor{blue}{rgb}{0.00,0.00,1.00}
\definecolor{green}{rgb}{0.4,1.00,0.0}
\definecolor{yellow}{rgb}{0.5,0.5,0.0}

\usepackage[pagebackref=true,breaklinks=true,colorlinks,bookmarks=false]{hyperref}
\wacvfinalcopy 

\begin{document}

\title{Dense Extreme Inception Network: Towards\\ a Robust CNN Model for Edge Detection}

\author{Xavier Soria$^{\dagger}$ \hspace{2cm}  Edgar Riba$^{\dagger}$ \hspace{2cm} Angel Sappa$^{\dagger,\ddagger}$ \\
$\dagger$ Computer Vision Center - Universitat Autonoma de Barcelona, Barcelona, Spain\\
$\ddagger$ Escuela Superior Polit\'ecnica del Litoral, Guayaquil, Ecuador\\
{\tt\small \{xsoria,eriba,asappa\}@cvc.uab.es}
}
\maketitle

\begin{abstract}
   This paper proposes a Deep Learning based edge detector, which is inspired on both HED (Holistically-Nested Edge Detection) and Xception networks. The proposed approach generates thin edge-maps that are plausible for human eyes; it can be used in any edge detection task without previous training or fine tuning process. As a second contribution, a large dataset with carefully annotated edges, has been generated. This dataset has been used for training the proposed approach as well the state-of-the-art algorithms for comparisons. Quantitative and qualitative evaluations have been performed on different benchmarks showing improvements with the proposed method when F-measure of ODS and OIS are considered.
\end{abstract}


\section{Introduction}
\label{sec:intro}

Edge detection is a recurrent task required for several classical computer vision processes (e.g., segmentation \protect \cite{zhang2016segmentation}, image recognition \cite{yang2002detectFace,shotton2008objectRec}), or even in the modern tasks such as image-to-image translation \cite{zhu2017cyclegan}, photo sketching \cite{lips2019photo-sketch} and so on. Moreover, in fields such as medical image analysis \cite{pourreza2017medImg} or remote sensing \cite{isikdogan2017remotesens} most of their heart activities require edge detectors. In spite of the large amount of work on edge detection, it still remains as an open problem with space for new contributions.

Since the Sobel operator \cite{sobel1972sobelmethod}, many edge detectors have been proposed \cite{oskoei2010surveyedge} and most of the techniques like Canny \cite{canny1987cannymethod} are still being used nowadays. Recently, in the era of Deep Learning (DL), Convolutional Neural Netwoks (CNN) based edge detectors like DeepEdge \cite{bertasius2015deepedge}, HED \cite{xie2017hed}, RCF \cite{liu2017rcf}, BDCN \cite{he2019edgeBDCN} among others, have been proposed. These models are capable of predicting an edge-map from a given image just like the low level based methods \cite{ziou1998edgeOverview}, with better performance. The success of these methods is mainly by the CCNs applied at different scales to a large set of images together with the training regularization techniques.

\begin{figure}
\begin{center}
\includegraphics[width=0.95\linewidth]{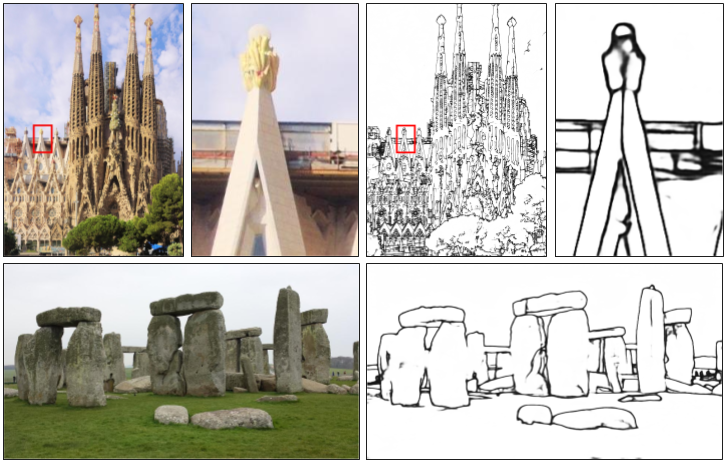}
\end{center}
  \caption{The edge-maps predictions from the proposed model in images acquired from internet.}
\label{fig:illustration}
\end{figure}


\begin{figure*}
\begin{center}
\includegraphics[width=0.95\linewidth]{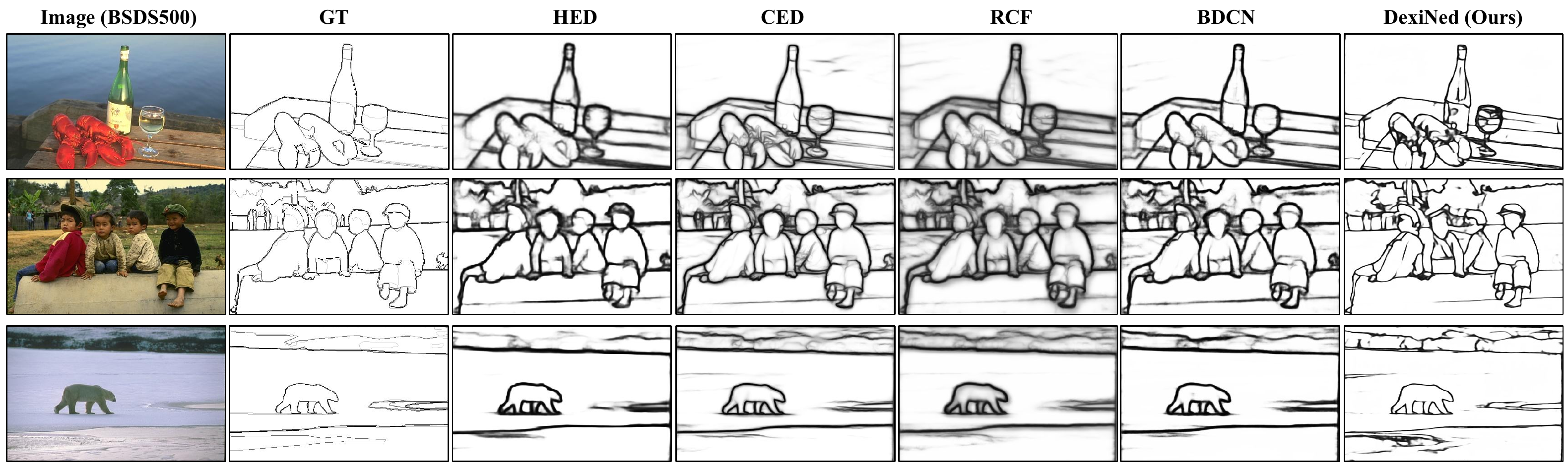}
\end{center}
   \caption{Edge-maps predicted from the state-of-the-art models and DexiNed on three BSDS500 \protect \cite{arbelaez2011bsds500} images. Note that DexiNed was just trained with BIPED, while all the others were trained on BSDS500.}
\label{fig:qual_comp}
\end{figure*}

Most of the aforementioned DL based approaches are trained on already existing boundary detection or object segmentation datasets \cite{martin2001bsds300,silberman2012NYUD, mottaghi2014PASCALcontext} to detect edges.
Even though most of the images on those datasets are well annotated, there are a few of them that contain missing edges, which difficult the training, thus the predicted edge-maps lost some edges in the images (see Fig. \ref{fig:illustration}). In the current work, those datasets are used just for qualitative comparisons due to the objective of the current work is edge detection (not objects' boundary/contour detection). The boundary/contour detection tasks, although related and some times assumed as a synonym task, are different since just objects' boundary/contour need to be detected, but not all edges present in the given image.

This manuscript aims to demonstrate the edge detection generalization from a DL model. In other words, the model is capable of being evaluated in other datasets for edge detection without being trained on those sets. To the best of our knowledge, the unique dataset for edge detection shared to the community is Multicue Dataset for Boundary Detection (MDBD---2016) \cite{mely2016multicue}, which although mainly generated for the boundary detection study, it contains a subset of images devoted for edge detection. Therefore, a new dataset has been collected to train the proposed edge detector. The main contributions in the paper are summarized as follow:

\begin{itemize}
    \item A dataset with carefully annotated edges has been generated and released to the community---BIPED: Barcelona Images for Perceptual Edge Detection.\footnote{Code $+$ dataset: \url{https://github.com/xavysp/DexiNed}}
    \item A robust CNN architecture for edge detection is proposed, referred to as DexiNed: Dense Extreme Inception Network for Edge Detection. The model has been trained from the scratch, without pretrained weights.
\end{itemize}

The rest of the paper is organized as follow. Section \ref{sec:rw} summarizes the most relevant and recent work on edge detection. Then, the proposed approach is described in Section \ref{sec:pa}. The experimental setup is presented in Section \ref{sec:exp}. Experimental results are then summarized in Section \ref{sec:res}; finally, conclusions and future work are given in Section \ref{sec:con}.

\section{Related Work}
\label{sec:rw}

\begin{figure*}[!ht]
\centering
\includegraphics[width=0.99\textwidth]{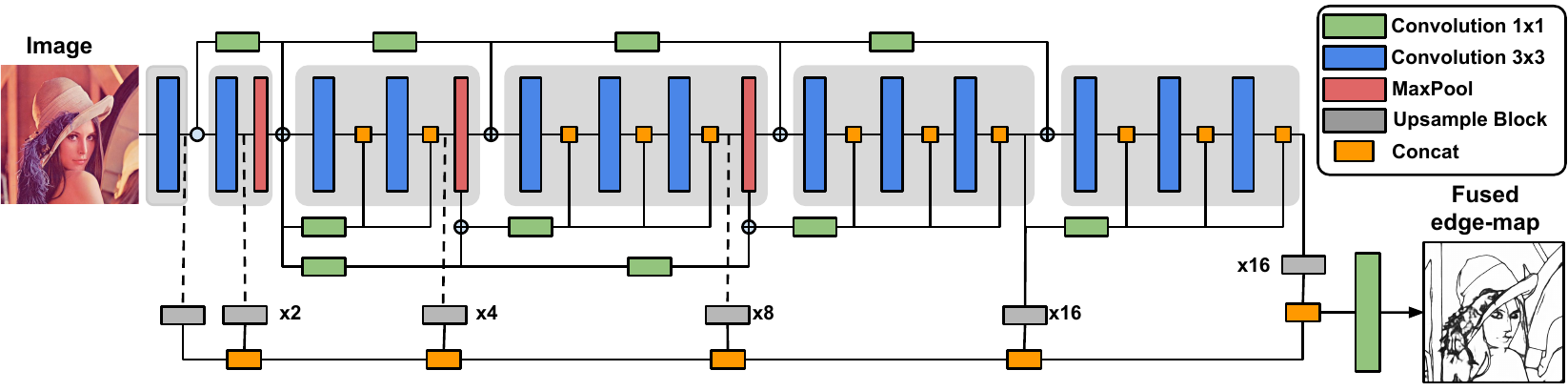}
\caption{Proposed architecture: Dense Extreme Inception Network, consists of an encoder composed by six main blocks (showed in light gray). The main blocks are connected between them through 1x1 convolutional blocks. Each of the main blocks is composed by sub-blocks that are densely interconnected by the output of the previous main block. The output from each of the main blocks is fed to an upsampling block that produces an intermediate edge-map in order to build a Scale Space Volume, which is used to compose a final fused edge-map. More details are given in Sec. \ref{sec:pa}.}
\label{fig:arch}
\end{figure*}

\begin{figure}
\centering
\includegraphics[width=0.35\textwidth]{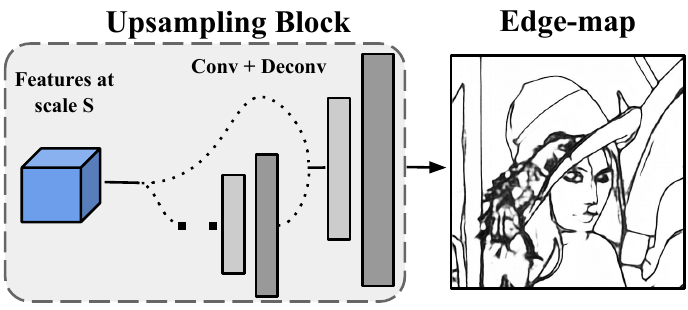}
\caption{Detail of the upsampling block that receives as input the learned features extracted from each of the main blocks. The features are fed into a stack of learned convolutional and transposed convolutional filters in order to extract an intermediate edge-map.}
\label{fig:upsampling-block}
\end{figure}

There are a large number of work on the edge detection literature, for a detailed review see \cite{ziou1998edgeOverview,Gong2018contourOverview}. According to the technique the given image is processed, proposed approaches can be categorized as: $i)$ Low level feature; $ii)$ Brain-biologically inspiration; $iii)$ Classical learning algorithms; $iv)$ Deep learning algorithms.

\textit{Low-level feature:} Most of the algorithms in this category generally follow a smooth process, which could be performed convolving the image with a Gaussian filter or manually performed kernels. A sample of such methods are \cite{canny1987cannymethod,schunck1987mMultiScaleF,perona1991mComEdges}. Since Canny \cite{canny1987cannymethod}, most of the nowadays methods use non-maximum suppression \cite{canny1983non-maximum} as the last process of edge detection.

\textit{Brain-biologically inspiration:} This kind of method started their research in the 60s of the last century analyzing the edge and contour formation in the vision systems of monkeys and cats \cite{daugman1985gaborFilter}. inspired on such a work, in \cite{grigorescu2003cid} the authors proposed a method based on simple cells and Gabor filters. Another study focused on boundary detection is presented in \cite{mely2016multicue}. This work proposes to use Gabor and derivative of Gaussian filters, considering three different filter sizes and machine learning classifiers. More recently, in \cite{yang2015SCO}, an orientation selective neuron is presented, by using first derivative of a Gaussian function. This work has been recently extended in \cite{Akbarinia2018SEDext} by modeling retina, simple cells even the cells from V2.

\textit{Classical learning algorithms:} These techniques are usually based on sparse representation learning \cite{mairal2008sparceModel}, dictionary learning \cite{xiaofeng2012Diclearn}, gPb (gradient descent) \cite{arbelaez2011bsds500} and structured forest \cite{dollar2015forests} (decision trees). At the time these approaches have been proposed, they outperformed state-of-the-art techniques based on low level processes reaching the best F-measure values in BSDS segmentation dataset \cite{arbelaez2011bsds500}. Although obtained results were acceptable in most of the cases, these techniques still have limitations in challenging scenarios.

\textit{Deep learning algorithms:} With the success of CNN, principally because of its result in \cite{krizhevsky2012alexnet}, many methods have been proposed \cite{ganin2014firstDLedge, bertasius2015deepedge, xie2017hed, liu2017rcf, wang2017ced}. In HED \cite{xie2017hed} for example, an architecture based on VGG16 \cite{simonyan2014vgg} and pre-trained with ImageNet dataset is proposed. The network generates edges from each convolutional block constructing a multi-scale learning architecture. The training process uses a modified cross entropy loss function for each predicted edge-maps. Using the same architecture as their backbone, \cite{liu2017rcf} and \cite{wang2017ced} have proposed improvements. While in \cite{liu2017rcf} every output is feed from each convolution from every block, in \cite{wang2017ced} a set of fusion backward process, with the data of each outputs, is performed. In general, most of the current DL based models use as their backbone the convolutional blocks of VGG16 architecture.

\section{Dense Extreme Inception Network for Edge Detection}
\label{sec:pa}

 \begin{figure*}[!ht]
 	\centering
 	\includegraphics[width=0.95\textwidth]{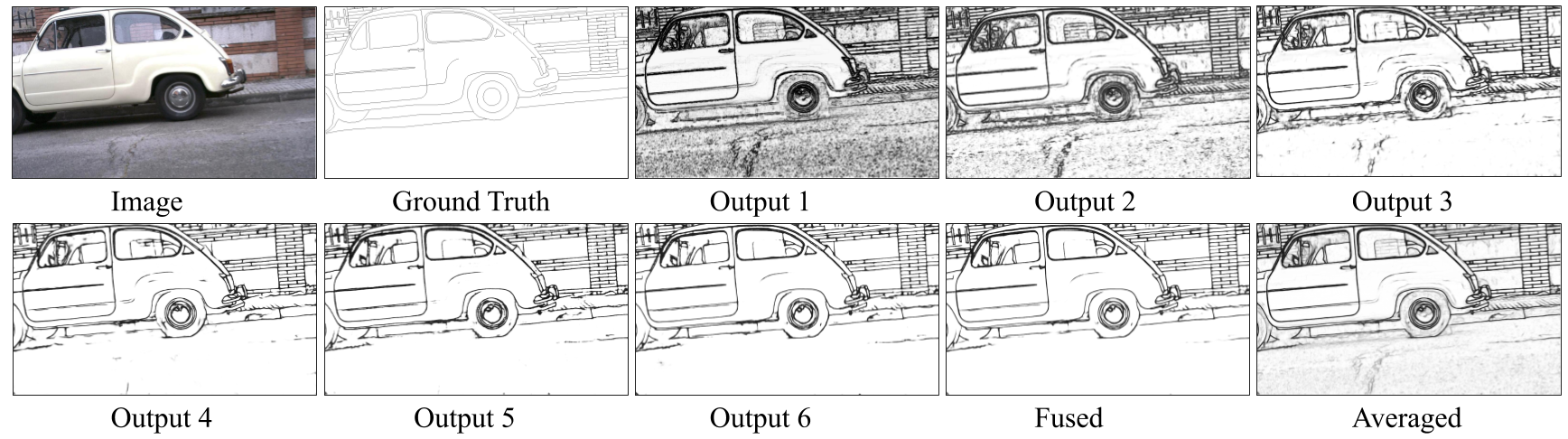}
 	\caption{Edge-maps from DexiNed in BIPED test dataset. The six outputs are delivered from the upsampling blocks, the \textit{fused} is the concatenation and fusion of those outputs and the \textit{averaged} is the average of all previous predictions.}
 	\label{fig:multiscaleresult}
 \end{figure*}

This section presents the architecture proposed for edge detection, termed DexiNed, which consists of a stack of learned filters that receive as input an image then predict an edge-map with the same resolution. DexiNed can be seen as two sub networks (see Figs. \ref{fig:arch} and \ref{fig:upsampling-block}): Dense extreme inception network (Dexi) and the up-sampling block (UB). While Dexi is fed with the RGB image, UB is fed with feature maps from each block of Dexi. The resulting network (DexiNed) generates thin edge-maps, avoiding missed edges in the deep layers. Note that even though without pre-trained data, the edges predicted from DexiNed are in most of the cases better than state-of-the-art results, see Fig. \ref{fig:illustration}.

\subsection{DexiNed Architecture}
\label{sec:pa-dexi}

The architecture is depicted in Fig. \ref{fig:arch}, it consists of an encoder with 6 main blocks inspired in the xception network \cite{chollet2017xception}. The network outputs feature maps at each of the main blocks to produce intermediate edge-maps using an upsampling block defined in Section \ref{sec:pa-upsampling}. All the edge-maps resulting from the upsampling blocks are concatenated to feed the stack of learned filters at the very end of the network and produce a fused edge-map. All six upsampling blocks do not share weights.

The blocks in blue consists of a stack of two convolutional layers with kernel size $3\times3$, followed by batch normalization and ReLU as the activation function (just the last convs in the last sub-blocks does not have such activation). The max-pool is set by $3\times3$ kernel and stride $2$. As the architecture follows the multi-scale learning, like in HED, an upsampling process (horizontal blocks in gray, Fig. \ref{fig:arch}) is followed (see details in Section \ref{sec:pa-upsampling}).

Even though DexiNed is inspired in xception, the similarity is just in the structure of the main blocks and connections. Major differences are detailed below:

\begin{itemize}
  \item While in xception separable convolutions are used, DexiNed uses standard convolutions.
  
  \item As the output is a 2D edge-map, there is "not exit flow", instead, another block at the end of block five has been added. This block has 256 filters and as in block 5 there is not maxpooling operator.
  
  \item In block 4 and  block 5, instead of 728 filters, 512 filters have been set. The separations of the main blocks are done with the blocks connections (rectangles in green) drawn on the top side of Fig. \ref{fig:arch}.
  
  \item Concerning to skip connections, in xception there is one kind of connection, while in DexiNed there are two type of connections, see rectangles in green on the top and bottom of Fig. \ref{fig:arch}.
\end{itemize}


Since many convolutions are performed, every deep block losses important edge features and just one main-connection is not sufficient, as highlighted in DeepEdge \cite{bertasius2015deepedge}, from the forth convolutional layer the edge feature loss is more chaotic. Therefore, since block $3$, the output of each sub-block is averaged with \textit{edge-connection} (orange squares in Fig. \ref{fig:arch}). These processes are inspired in ResNet \cite{he2016resnet} and RDN \cite{zhang2018densenet} with the following notes: $i)$ as shown in Fig. \ref{fig:arch}, after the max-pooling operation and before summation with the main-connection, the edge-connection is set to average each sub-blocks output (see rectangles in green, bottom side); $ii)$ from the max-pool, block $2$, edge-connections feed sub-blocks in block $3$, $4$ and $5$, however, the sub-blocks in $6$ are feed just from block 5 output.

\subsection{Upsampling Block}
\label{sec:pa-upsampling}

DexiNed has been designed to produce thin edges in order to enhance the visualization of predicted edge-maps. One of the key component of DexiNed for the edge thinning is the upsampling block, as appreciated in Fig. \ref{fig:arch}, each output from the Dexi blocks feeds the UB. The UB consists of the conditional stacked sub-blocks. Each sub-block has 2 layers, one convolutional and the other deconvolutional; there are two types of sub-blocks. The first sub-block (sub-block1) is feed from Dexi or sub-block2; it is only used when the scale difference between the feature map and the ground truth is equal to 2. The other sub-block (sub-block2), is considered when the difference is greater than 2. This sub-block is iterated till the feature map scale reaches 2 with respect to the GT. The sub-block1 is set as follow: kernel size of the conv layer $1\times1$; followed by a ReLU activation function; kernel size of the deconv layer or transpose convolution $s \times s$, where $s$ is the input feature map scale level; both layers return one filter and the last one gives a feature map with the same size as the GT. The last conv layer does not have activation function. The sub-block2 is set similar to sub-block1 with just one difference in the number of filters, which is 16 instead of 1 in sub-block1. For example, the output feature maps from block 6 in Dexi has the scale of $16$, there will be three iterations in the sub-block2 before fed the sub-block1. The upsampling process of the second layer from the sub-blocks can be performed by bi-linear interpolation, sub-pixel convolution and transpose convolution, see Sec. \ref{sec:res} for details.

\subsection{Loss Functions}
\label{sec:pa-loss}

DexiNed could be summarized as a regression function $\eth$, that is, $\hat{Y}$ = $\eth(X,Y)$, where $X$ is an input image, $Y$ is its respective ground truth, and $\hat{Y}$ is a set of predicted edge maps. $\hat{Y} = [\hat{y}_{1},\hat{y}_{2},...,\hat{y}_{N}]$, where $\hat{y}_i$ has the same size as $Y$, and $N$ is the number of outputs from each upsampling block (horizontal rectangles in gray, Fig. \ref{fig:arch});
$\hat{y}_{N}$ is the result from the last fusion layer $f$ $(\hat{y}_{N}=\hat{y}_{f}$). Then, as the model is deep supervised, it uses the same loss as \cite{xie2017hed} (weighted cross-entropy), which is tackled as follow:

\begin{equation}
\centering
\begin{split}
\mathcal{o}^{n}(W,w^{n}) &=- \beta \sum_{j\in{Y^+}} \log{\sigma(y_j =1|X;W,w^n)}\\
&-(1-\beta) \sum_{j \in{Y^-}} \log{\sigma(y_j =0|X;W,w^n)},
\end{split}
\label{eq:sin-loss}
\end{equation}

\noindent then,

\begin{equation}
\centering
\mathcal{L}(W,w)=\sum_{n=1}^{N}\delta^n\times\mathcal{o}^{n}(W,w^{n}),
\label{eq:sum-loss}
\end{equation}

\noindent where $W$ is the collection of all network parameters and $w$ is the $n$ corresponding parameter, $\delta$ is a weight for each scale level. $\beta$ = $|Y^-|/|Y^+ + Y^-|$ and $(1-\beta)$=$|Y^+|/|Y^+ + Y^-|$ ($|Y^-|$, $|Y^+|$ denote the edge and non-edge in the ground truth). See Section \ref{sub:impl-notes} for hyper-parameters and optimizer details for the regularization in the training process.

\section{Experimental Setup}
\label{sec:exp}
This section presents details on the datasets used for evaluating the proposed model, in particular the dataset and annotations (BIPED) generated for an accurate training of the proposed DexiNed. Additionally, details on the evaluation metrics and network's parameters are provided.

\subsection{Barcelona Images for Perceptual Edge Detection (BIPED)}
\label{sub:BIPED}

The other contributions of the paper is a carefully annotated  edge dataset. It contains 250 outdoor images of 1280$\times$720 pixels each. These images have been carefully annotated by experts on the computer vision field, hence no redundancy has been considered. In spite of that, all results have been cross-checked in order to correct possible mistakes or wrong edges. This dataset is publicly available as a benchmark for evaluating edge detection algorithms. The generation of this dataset is motivated by the lack of edge detection datasets, actually, there is just one dataset publicly available for the edge detection task (MDBD \cite{mely2016multicue}). Edges in MDBM dataset have been generated by different subjects, but have not been validated, hence, in some cases, the edges correspond to wrong annotations. Some examples of these missed or wrong edges can be appreciated in the ground truths presented in Fig. \ref{fig:diferentedatasets}; hence, edge detector algorithms that obtain these missed edges are penalized during the evaluation. The level of details of the dataset annotated in the current work can be appreciated looking at the GT, see Figs. \ref{fig:multiscaleresult} and \ref{fig:illustrationcomparisons}. In order to do a fair comparison between the different state-of-the-art approaches proposed in the literature, BIPED dataset has been used for training those approaches, which have been later on evaluated in ODS, OIS, and AP. From the BIPED dataset, 50 images have been randomly selected for testing and the remainders 200 for training and validation. In order to increase the number of training images a \textbf{data augmentation process }has been performed as follow: i) as BIPED data are in high resolution they are split up in the half of image width size; ii) similarly to HED, each of the resulting images is rotated by 15 different angles and crop by the inner oriented rectangle; iii) the images are horizontally flip; and finally iv) two gamma corrections have been applied (0.3030, 0.6060). This augmentation process resulted in 288 images per each 200 images.

\subsection{Test Datasets}
\label{sub:test-data}

The datasets used to evaluate the performance of DexiNed are summarized bellow. There is just one dataset intended for edged detection MDBD \cite{mely2016multicue}, while the remainders are for objects' contour/boundary extraction/segmentation: CID \cite{grigorescu2003cid}, BSDS \cite{martin2001bsds300, arbelaez2011bsds500}, NYUD \cite{silberman2012NYUD} and PASCAL \cite{mottaghi2014PASCALcontext}.

\textit{MDBD:} The Multicue Dataset for Boundary Detection  has been intended for the purpose of psychophysical studies on object boundary detection in natural scenes, from the early vision system. The dataset is composed of short binocular video sequences of natural scenes \cite{mely2016multicue}, containing 100 scenes in high definition ($1280\times720$). Each scene has 5 boundary annotations and 6 edge annotations. From the given dataset 80 images are used for training and the remainders 20 for testing \cite{mely2016multicue}. In the current work, DexiNed has been evaluated using the first 20 images (the sub set for edge detection).

\textit{CID:} This dataset has been presented in \cite{grigorescu2003cid}, a brain-biologically inspired edge detector technique. The main limitation of this dataset is that it just contains a set of 40 images with their respective ground truth edges. This dataset highlight that in addition to the edges the ground truth map contains contours of object. In this case the DexiNed has been evaluated with the whole CID data.

\textit{BSDS:} Berkeley Segmentation Dataset, consists of 200 new test images \cite{arbelaez2011bsds500} additional to the 300 images contained in BSDS300 \cite{martin2001bsds300}. In previous publications, the BSDS300 is split up into 200 images for training and 100 images for testing. Currently, the 300 images from BSDS300 are used for training and validation, while the remainders 200 images are used for testing. Every image in BSDS is annotated at least by $6$ annotators; this dataset is mainly intended for image segmentation and boundary detection. In the current work both datasets are evaluated BSDS500 (200 test images) and BSDS300 (100 test images).

\textit{NYUD:} New York University Dataset is a set of 1449 RGBD images that contains 464 indoor scenarios, intended for segmentation purposes. This dataset is split up by \cite{Gupta_2013NYUDsplit} into three subsets---i.e., training, validation and testing sets. The testing set contains 654 images, while the remainders images are used for training and validation purposes. In the current work, although the proposed model was not trained with this dataset, the testing set has been selected for evaluating the proposed DexiNed. 

\textit{PASCAL:} The Pascal-Context \cite{mottaghi2014PASCALcontext} is a popular dataset in segmentation; currently most of major DL methods for edge detection use this dataset for training and testing, both for edge and boundary detection purposes. This dataset contains 11530 annotated images, about $5\%$ of them (505 images) have been considered for testing DexiNed.

\subsection{Evaluation Metrics}
\label{sub:em}

The evaluation of an edge detector has been well defined since the pioneer work presented in \cite{ziou1998edgeOverview}.
Since BIPED has annotated edge-maps as GT, three evaluation metrics widely used in the community have been considered: fixed contour threshold (ODS), per-image best threshold (OIS), and average precision (AP). The F-measure (F) \cite{arbelaez2011bsds500} of ODS and OIS, will be considered, where $F=\frac{2\times Precision \times Recall}{Precision + Recall}$.


\subsection{Implementation Notes}
\label{sub:impl-notes}

The implementation is performed in TensorFlow \cite{abadi2016tensorflow}. The model converges after 150k iterations with a batch size of 8 using Adam optimizer and learning rate of $10^{-4}$. The training process takes around 2 days in a TITAN X GPU with color images of size 400x400 as input. The weights for fusion layer are initialized as: $\frac{1}{N-1}$ (see Sec. \ref{sec:pa-loss} for $N$). After a hyperparameter search to reduce the number of parameters, best performance was obtained using kernel sizes of $3\times3$, $1\times1$ and $s\times s$ on the different convolutional layers of Dixe and UB.


\section{Experimental Results}
\label{sec:res}

\begin{figure*}
\begin{tabular}{ccc}
        \includegraphics[width=0.30\textwidth]{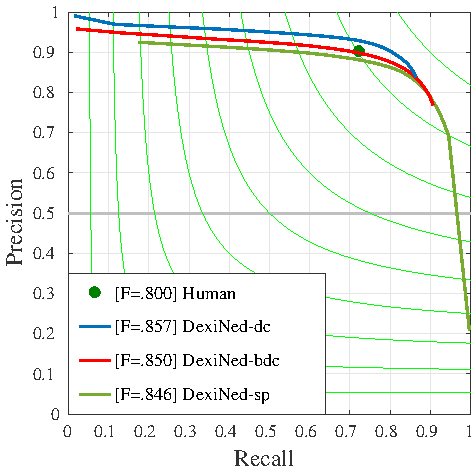} &
        \includegraphics[width=0.30\textwidth]{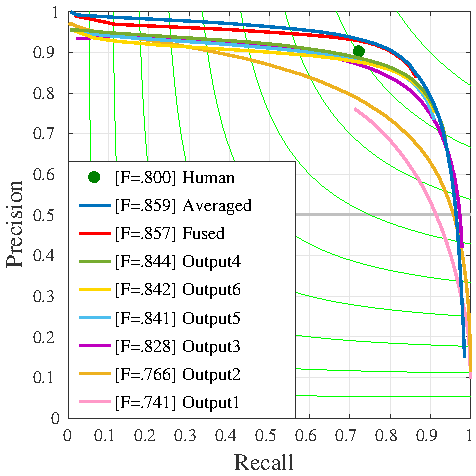} &
        \includegraphics[width=0.30\textwidth]{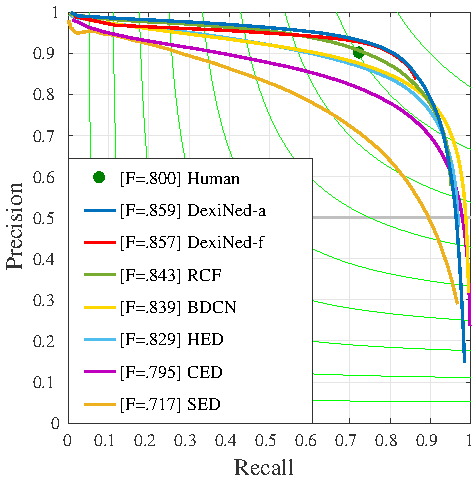} \\
        (a) & (b) & (c)\\
\end{tabular}      
   \caption{Precision/recall curves on BIPED dataset. (a) DexiNed upsampling versions. (b) The outputs of DexiNed in testing stage, the 8 outputs are considered. (c) DexiNed comparison with other DL based edge detectors.}
    \label{fig:curves}
\end{figure*}

\begin{table}\small
    \centering

\begin{tabular}{ll}
\centering
\setlength\tabcolsep{0.7pt} 
\begin{tabular}{c|c|c|c}
    \hline
  Outputs& ODS& OIS& AP\\
\hline\hline
Output 1 ($\hat{y}_{1}$) & .741&.760&.162 \\
Output 2 ($\hat{y}_{2}$)& .766&.803&.817 \\
Output 3 ($\hat{y}_{3}$)& .828&.846&.838 \\
Output 4 ($\hat{y}_{4}$)& .844&.858&.843\\
Output 5 ($\hat{y}_{5}$)& .841&.8530&.776\\
Output 6 ($\hat{y}_{6}$)& .842&.852&.805\\
Fused ($\hat{y}_{f}$)& .857&.861&.805\\
Averaged &\textbf{.859}&\textbf{.865}&\textbf{.905}\\
\hline
\end{tabular}
&\hspace{-0.2cm}
\setlength\tabcolsep{1.7pt} 
\begin{tabular}{c|c|c|c}
    \hline
  Methods & ODS& OIS& AP\\
\hline\hline
SED\cite{Akbarinia2018SEDext} &.717&.731&.756 \\
HED\cite{xie2017hed} & .829&.847&.869 \\
CED\cite{wang2017ced} & .795&.815&.830\\
RCF\cite{liu2019RCFext} & .843&.859&.882\\
BDCN\cite{he2019edgeBDCN} & .839&.854&.887\\
DexiNed-f & .857&.861&.805\\
DexiNed-a &\textbf{.859}&\textbf{.867}&\textbf{.905}\\
\hline
\end{tabular}
\end{tabular}\\
\vspace{0.1cm}
\hspace{1cm} (a) \hspace{3.5cm} (b)\\
    \caption{(a) Quantitative evaluation of the 8 predictions of DexiNed on BIPED test dataset. $(b)$ Comparisons between the state-of-the-art methods trained and evaluated with BIPED.}
    \label{tab:BIPED}
\end{table}

This section presents quantitative and qualitative evaluations conducted by the metrics presented in Sec. \ref{sec:exp}. Since the proposed DL architecture demands several experiments to be validated, DexiNed has been carefully tuned till reach its final version.

\begin{table}
\begin{center}

\begin{tabular}{c|c|c|c|c}
\hline
Dataset& Methods& ODS & OIS & AP\\
\hline\hline

\multicolumn{5}{c}{Edge detection dataset}\\
\hline
MDBD\cite{mely2016multicue} &HED\cite{xie2017hed}&.851&\textbf{.864}&.890 \\
 &RCF\cite{liu2017rcf}&.857&.862&-\\
 &DexiNed-f&.837&.837&.751\\
 &DexiNed-a&.\textbf{859}&\textbf{.864}&\textbf{.917}\\
\hline
\multicolumn{5}{c}{Contour/boundary detection/segmentation  datasets}\\
\hline
 \hline
CID\cite{grigorescu2003cid} &SCO\cite{yang2015SCO}&.58&.64&.61\\
 &SED\cite{Akbarinia2018SEDext}&\textbf{.65}&\textbf{.69}&.68 \\
 &DexiNed-f&.65&.67&.59\\
  &DexiNed-a&\textbf{.65}&\textbf{.69}&\textbf{.71}\\
 \hline
 BSDS300\cite{martin2001bsds300}&gPb\cite{arbelaez2011bsds500}&.700&.720&.660\\
 &SED\cite{Akbarinia2018SEDext}&.69&.71&.71\\
 &DexiNed-f&.707&.723&.52\\
  &DexiNed-a&\textbf{.709}&\textbf{.726}&\textbf{.738}\\
 \hline
  BSDS500\cite{arbelaez2011bsds500}&HED\cite{xie2017hed}&.790&.808&.811\\
 &RCF\cite{liu2017rcf}&\textbf{.806}&\textbf{.823}&-\\
 &CED\cite{wang2017ced}&.803&.820&\textbf{.871}\\
 &SED\cite{Akbarinia2018SEDext}&.710&.740&.740\\
 &DexiNed-f&.729&.745&.583\\
  &DexiNed-a&.728&.745&.689\\
  \hline
NYUD\cite{silberman2012NYUD}&gPb\cite{arbelaez2011bsds500}&.632&.661&.562\\
&HED\cite{xie2017hed}&.720&\textbf{.761}&\textbf{.786} \\
 &RCF\cite{liu2017rcf}&\textbf{.743}&.757&-\\
 &DexiNed-f&.658&.674&.556\\
  &DexiNed-a&.602&.615&.490\\
 \hline
PASCAL\cite{mottaghi2014PASCALcontext}&CED\cite{wang2017ced}&\textbf{.726}&\textbf{.750}&\textbf{.778} \\
 &HED\cite{xie2017hed}&.584&.592&.443\\
 &DexiNed-f&.431&.458&.274\\
  &DexiNed-a&.475&.497&.329\\

\hline
\end{tabular}
\end{center}
\caption{Quantitative results of \textbf{DexiNed trained on BIPED} and \textbf{the state-o-the-art methods trained with the corresponding datasets} (values from other approaches come from the corresponding publications).}
\label{tab:alldata}
\end{table}

\subsection{Quantitative Results}

Firstly, in order to select the upsampling process that achieves the best result, an empiric evaluation has been performed, see Fig. \ref{fig:curves}(a). The evaluation consists in conducting the same experiments by using the three upsampling methods; \textbf{DexiNed-bdc} refers to upsampling performed by a transpose convolution initialized with a bi-linear kernel; \textbf{DexiNed-dc} uses transpose convolution with trainable kernels; and \textbf{DexiNed-sp} uses subpixel convolution. According to F-measure, the three versions of DexiNed get the similar results, however, when analyzing the curves in Fig. \ref{fig:curves}(a), a small difference in the performance of DexiNed-dc appears. As a conclusion, the DexiNed-dc upsampling strategy is selected; from now on, all the evaluations performed on this section are obtained using a DexiNed-dc upsampling; for simplicity of notation just the term DexiNed is used instead of DexiNed-dc.

Figure \ref{fig:curves}(b) and Table \ref{tab:BIPED}(a) present the quantitative results reached from each DexiNed edge-map prediction. The results from the eight predicted edge-maps are depicted, the best quantitative results, corresponding to the fused (DexiNed-f) and averaged (DexiNed-a) edge-maps are selected for the comparisons. Similarly to \cite{xie2017hed} the averaged of all predictions (DexiNed-a) gets the best results in the three evaluation metrics, followed by the prediction generated in the fusion layer. Note that the edge-maps predicted from the block 2 till the 6 get similar results to DexiNed-f, this is due to the fact of the proposed skip-connections. For a qualitative illustration, Fig. \ref{fig:multiscaleresult} presents all edge-maps predicted from the proposed architecture. Qualitatively, the result from DexiNed-f is considerably better than the one from DexiNed-a (see illustration in Fig. \ref{fig:multiscaleresult}). However, according to Table \ref{tab:BIPED}(a), DexiNed-a produces slightly better quantitative results than DexiNed-f. As a conclusion both approaches (fused and averaged) reach similar results; through this manuscript whenever the term DexiNed is used it corresponds to DexiNed-f.

 \begin{figure*}
 	\centering
 	\includegraphics[width=0.93\textwidth]{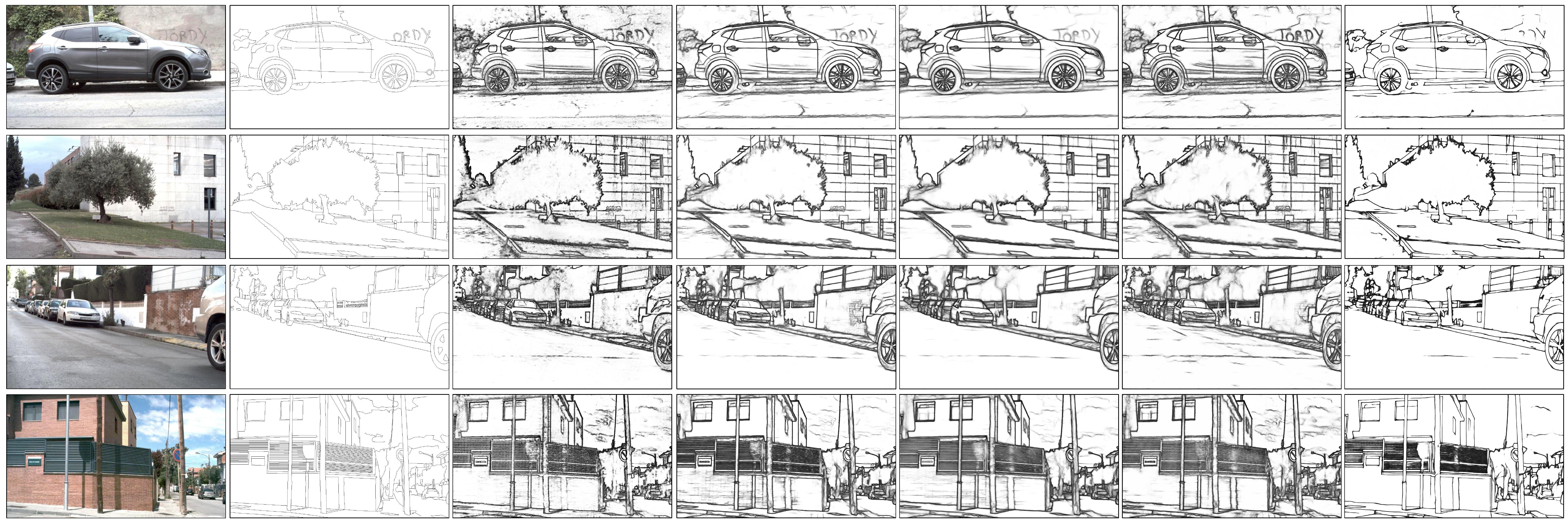}\\
 	\hspace{0.3cm}\textit{Image} \hspace{1.3cm} \textit{GT} \hspace{1.3cm} \textit{CED \cite{wang2017ced}} \hspace{1.2cm}  \textit{HED \cite{xie2017hed}} \hspace{0.9cm}  \textit{RCF \cite{liu2017rcf}} \hspace{0.7cm} 
 	\textit{BDCN \cite{he2019edgeBDCN}} \hspace{0.6cm} \textit{DexiNed}\\
 	\caption{Results from different edge detection algorithms trained and evaluated in BIPED dataset.}
 	\label{fig:illustrationcomparisons}
 \end{figure*}
 
 \begin{figure*}
 	\centering
 	\includegraphics[width=0.93\textwidth]{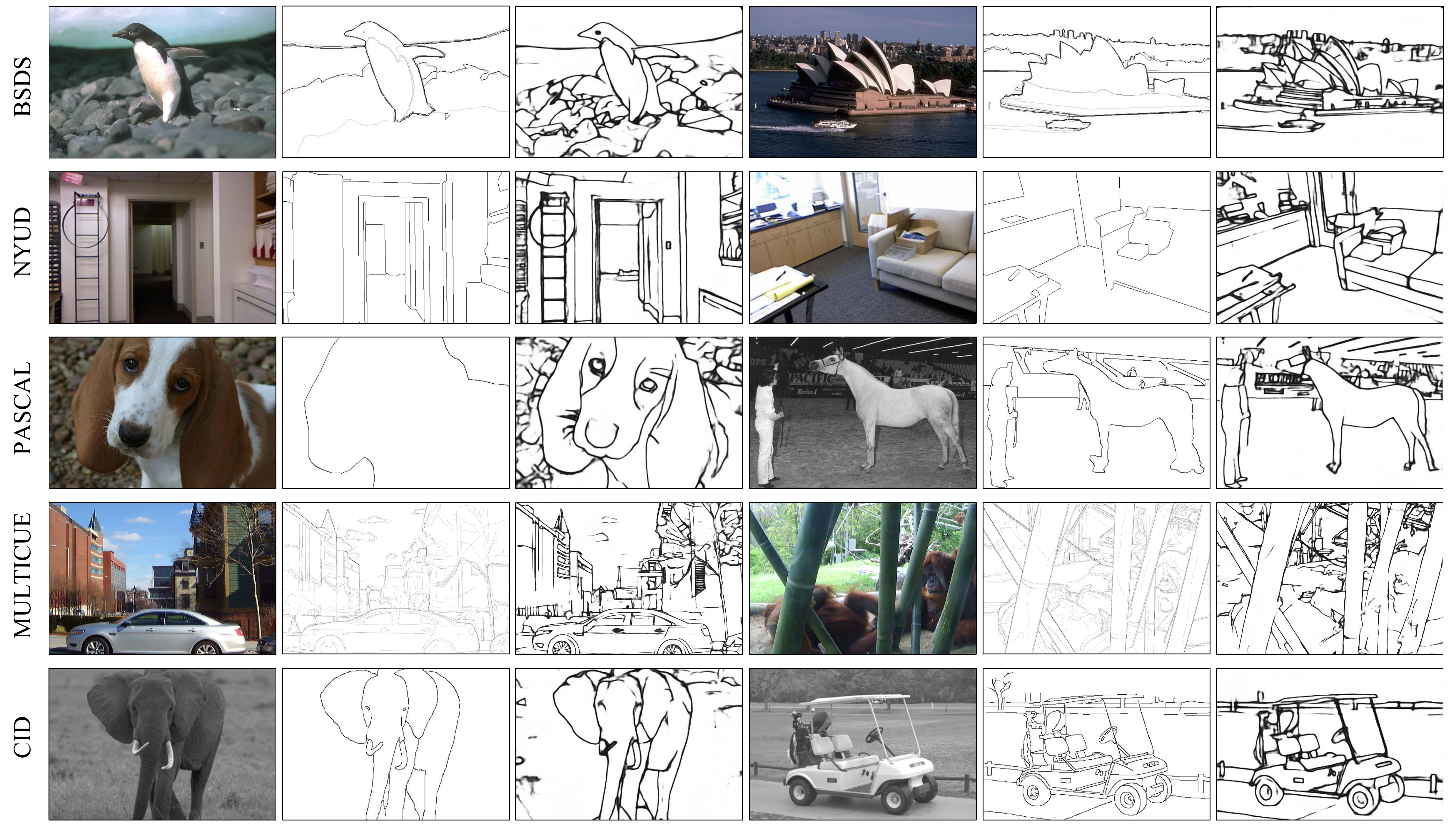}\\ 
 	\hspace{1cm}\textit{Image} \hspace{1.6cm} \textit{GT} \hspace{1.6cm} \textit{DexiNed} \hspace{1.6cm}  \textit{Image} \hspace{1.6cm}  \textit{GT} \hspace{1.6cm}  \textit{DexiNed}\\
 	\caption{Results from the proposed approach using different datasets (note that DexiNed has been trained just with BIPED).}
 	\label{fig:diferentedatasets}
 \end{figure*}
Table \ref{tab:BIPED}(b) presents a comparison between the DexiNed and the state-of-the-art techniques on edge and boundary detection. In all the cases BIPED dataset has been considered, both for training and evaluating the DL based models (i.e., HED \cite{shen2015deepcontour}, RCF \cite{liu2017rcf}, CED \cite{wang2017ced}) and BDCN \cite{he2019edgeBDCN}, the training process for each model took about two days. As can be appreciated from Table \ref{tab:BIPED}(b), DexiNed-a reaches the best results in all evaluation metrics. Actually both, DexiNed-a and DexiNed-f obtain the best results in almost all evaluation metrics. The F-measure obtained by comparing these approaches is presented in Fig. \ref{fig:curves}(c); it can be appreciated how for Recall above 75\% DexiNed gets the best results. Illustrations of the edges obtained with DexiNed and the state-of-the-art techniques are depicted in Figure \ref{fig:illustrationcomparisons}, just for four images from the BIPED dataset. As it can be appreciated, although RCF and BDCN obtain similar quantitative results than DexiNed, which were the second best ranked algorithms in Table \ref{tab:BIPED}(b), DexiNed predicts qualitative better results. Note that the proposed approach was trained from scratch without pre-trained weights.

The main objective of DexiNed is to get a precise edge-map from every dataset (RGB or Grayscale). Therefore, all the datasets presented in Sec. \ref{sub:test-data} have been considered, split up into two categories for a fair analysis; one for \textbf{edge detection} and the others for \textbf{contour/boundary detection/segmentation}. Results of edge-maps obtained with state-of-the-art methods are presented in Table \ref{tab:alldata}. It should be noted that for each dataset the methods compared with DexiNed have been trained using images from that dataset, while DexiNed is trained just once with BIPED. It can be appreciated that DexiNed obtains the best performance in the MDBD dataset. It should be noted that DexiNed is evaluated in CID and BSDS300, even though these datasets contain a few images, which are not enough for training other approaches (e.g., HED, RCF, CED). Regarding BSDS500, NYUD and PASCAL, DexiNed does not reach the best results since these datasets have not been intended for edge detection, hence the evaluation metrics penalize edges detected by DexiNed. To highlight this situation,  Fig. \ref{fig:diferentedatasets} depicts results from Table \ref{tab:alldata}. Two samples from each dataset are considered. They are selected according to the best and worst F measure. Therefore, as shown in Fig. \ref{fig:diferentedatasets}, when the image is fully annotated the score reaches around 100\%, otherwise it reaches less than 50\%.


\subsection{Qualitative Results}
\label{sub:qualr}

As highlighted in previous section, when the deep learning based edge detection approaches are evaluated in datasets intended for objects' boundary detection or objects segmentation, the results will be penalized. To support this claim, we present in Fig. \ref{fig:diferentedatasets} two predictions (the best and the worst results according to F-measure) from all datasets used for evaluating the proposed approach (except BIPED that has been used for training). The F-measure obtained in the three most used datasets (i.e., BSDS500, BSDS300 and NYUD) reaches over 80$\%$ in those cases where images are fully annotated; otherwise, the F-measure reaches about 30$\%$. However, when the edge dataset (MDBD \cite{mely2016multicue}) is considered the worst F-measure reaches over 75$\%$. As a conclusion, it should be stated that edge detection and contour/boundary detection are different problems that need to be tackled separately when a DL based model is considered.



\section{Conclusions}
\label{sec:con}

A deep structured model (DexiNed) for image's edge detection is proposed. Up to our knowledge, it is the first DL based approach able to generate thin edge-maps. A large experimental results and comparisons with state-of-the-art approaches is provided showing the validity of DexiNed. Even though DexiNed is trained just one time (with BIPED) it outperforms the state-of-the-art approaches when evaluated in other edge oriented datasets. A carefully annotated dataset for edge detection has been generated and is shared to the community. Future work will be focused on tackling the contour and boundary detection problems by using the proposed architecture and approach.

\section*{Acknowledgment}

This work has been partially supported by: the Spanish Government under Project TIN2017-89723-P; the ``CERCA Programme / Generalitat de Catalunya" and the ESPOL project PRAIM (FIEC-09-2015). The authors gratefully acknowledge the support of the CYTED Network: ``Ibero-American Thematic Network on ICT Applications for Smart Cities'' (REF-518RT0559) and the NVIDIA Corporation with the donation of the Titan Xp GPU used for this research. Xavier Soria has been supported by Ecuador government institution SENESCYT under a scholarship contract 2015-AR3R7694.

\balance

{\small
\bibliographystyle{ieee}
\bibliography{main.bib}
}

\end{document}